\documentclass[11pt,letterpaper,logo]{googledeepmind}

\usepackage{adjustbox}
\usepackage{threeparttable}
\usepackage{multirow}
\usepackage{algorithm}
\usepackage{algpseudocode}
\usepackage{cite}
\usepackage{flafter}
\usepackage{etoolbox}
\apptocmd{\thebibliography}{\setlength{\itemsep}{3pt}\setlength{\parsep}{0pt}}{}{}

\definecolor{OursRow}{RGB}{240,240,240}
\newcommand{\cmark}{\ding{51}}
\newcommand{\xmark}{\ding{55}}
\newcolumntype{Y}{>{\centering\arraybackslash}X}

\hypersetup{colorlinks=true,citecolor=magenta,
  pdftitle={AsymEEF: Bimanual Manipulation with Asymmetric End Effectors},
  pdfauthor={Xiaokang Sun, Ningyue Wang, Kunqi Xu, Guangqi Jiang, Jianglong Ye, Yong Wu, Xueyan Zou}}
\pdftrailerid{redacted}

\fancypagestyle{firststyle}{%
  \fancyhf{}
  \fancyhead[L]{\raisebox{-2pt}{\includegraphics[height=16pt,trim=0 0 176.73bp 0,clip]{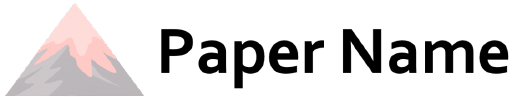}}\hspace{4pt}{\sffamily\bfseries AsymEEF}}
  \fancyfoot[R]{\footerfont\thepage}
}
\titlespacing*{\section}{0pt}{2.5ex plus 2pt minus 1pt}{5pt}
\titlespacing*{\subsection}{0pt}{2ex plus 2pt minus 1pt}{2pt}
\titlespacing*{\subsubsection}{0pt}{1.8ex plus 1.5pt minus 1pt}{2pt}
\renewcommand{\absfont}{\bfseries\linespread{1.15}\fontsize{10}{12}\selectfont}
\patchcmd{\maketitle}{\vskip11pt}{\vskip8pt}{}{}
\patchcmd{\maketitle}{\vskip20pt}{\vskip12pt}{}{}

\title{\textit{AsymEEF}: Bimanual Manipulation with Asymmetric End Effectors}
\author[1,2,*]{Xiaokang Sun}
\author[2,*]{Ningyue Wang}
\author[2]{Kunqi Xu}
\author[2]{Guangqi Jiang}
\author[3]{Jianglong Ye}
\author[2]{Yong Wu}
\author[2]{Xueyan Zou}
\affil[1]{Georgia Institute of Technology}
\affil[2]{Tsinghua University}
\affil[3]{University of California San Diego}
\affil[*]{Equal contribution}
\begin{abstract}
Teleoperated demonstrations are costly to collect and rarely reusable across end-effectors with different kinematics and action spaces. This burden grows in bimanual manipulation, where each end-effector pair may require new data. We present AsymEEF, a framework for executable cross-end-effector demonstration generation and unified bimanual policy learning. Given a small source dataset, AsymEEF transfers demonstrations to heterogeneous combinations of \mbox{two-,} three-, and five-fingered end-effectors without additional teleoperation. Kinematic teacher anchors provide geometric supervision, while sparse real-world X-Calib anchors correct discrepancies in grasp execution. A shared encoder maps wrist poses and actuator states to a morphology-agnostic latent space, and a morphology-conditioned decoder reconstructs target commands. Generated trajectories undergo replay filtering for task-level executability. The representation also supports a unified bimanual visuomotor policy trained across multiple pairs. Across four simulated and six real-world tasks, AsymEEF improves replay success over adapted transfer baselines, achieves a mean replay-valid expansion gain of 16.6$\times$ in simulation, and is evaluated through downstream policy training on real robots. Project website: \url{https://asymeef.vercel.app/}.

\end{abstract}

\begin{document}
\maketitle

\par\vspace{8pt plus 1fill}
\noindent\begin{minipage}{\linewidth}
    \centering
    \includegraphics[width=\linewidth]{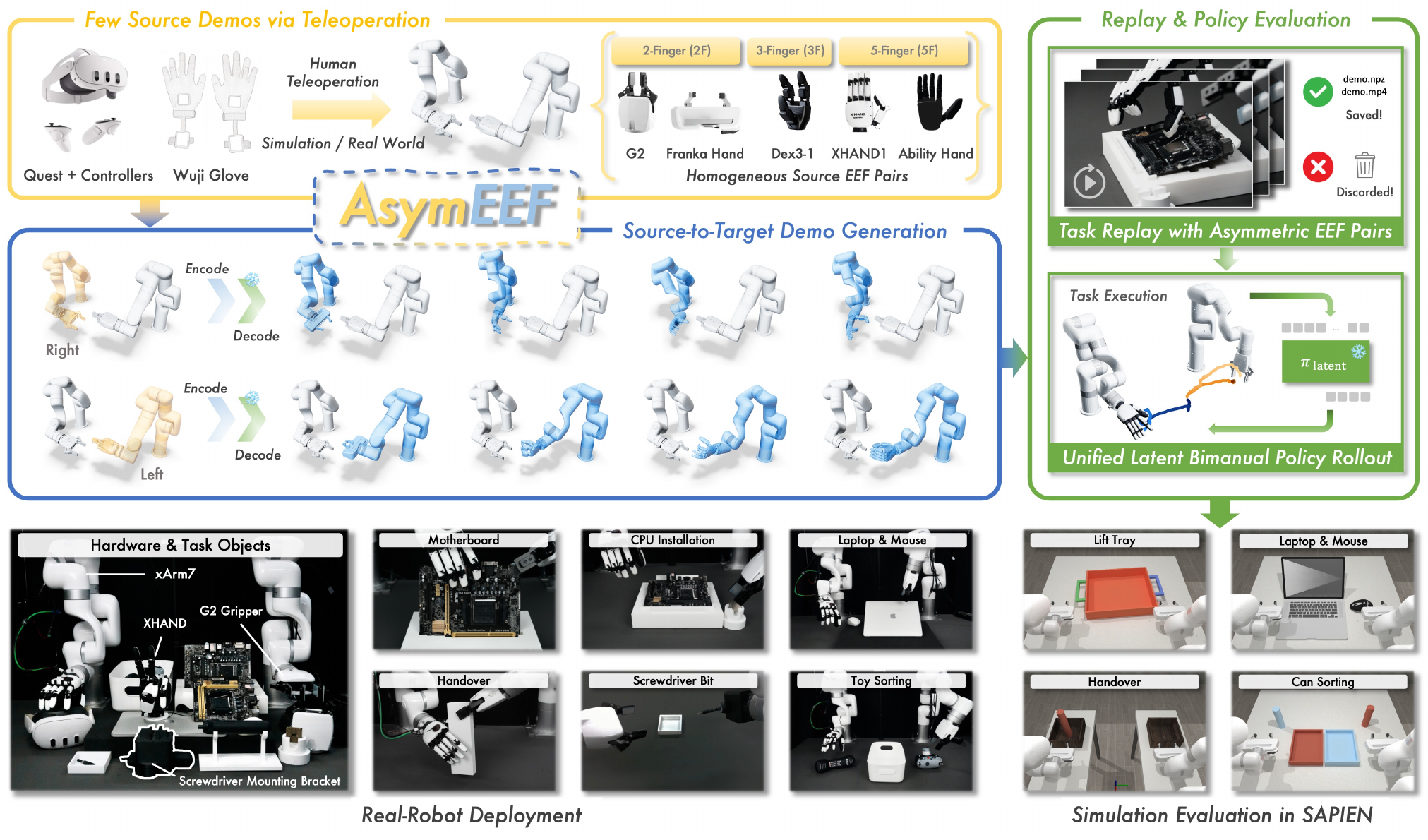}
    \captionsetup{hypcap=false}\captionof{figure}{Overview. AsymEEF converts a small set of teleoperated demonstrations from homogeneous end-effector pairs into trajectories for bimanual robots with heterogeneous end-effectors, including grippers and dexterous hands. Generated data are validated through strict task replay and used to train a unified latent bimanual visuomotor policy. We evaluate the framework across diverse contact-rich manipulation tasks in simulation and on real robots.}
    \label{fig:teaser}
\end{minipage}\par
\pagebreak

\section{Introduction}

High-quality demonstrations are central to imitation learning for robotic manipulation, yet collecting them through teleoperation is time-consuming and costly~\cite{mandlekar2023mimicgen,xue2025demogen,jiang2025dexmimicgen}. Reusing these demonstrations across end effectors is challenging because the recorded actions are tied to the source hardware's action space, while the resulting motions and interactions depend on its kinematics and contact behavior~\cite{zheng2025uniact,bauer2025latentdiff,mu2026one,jiang2026cross,wu2026cei}. Directly transferring these actions to another end effector may therefore fail to reproduce the demonstrated interaction. This motivates data-generation methods that not only expand datasets across object types and positions but also support demonstration reuse across end effectors.

We study transfer among morphologically and kinematically distinct two-, three-, and five-fingered end effectors, which we collectively call \textit{hands}. Our objective is to reuse demonstrations from limited source hand types across other hands and bimanual pairs. Source hands may be selected for cost, usability, or existing teleoperation support. For example, when the target hands support the required interactions, demonstrations collected with $A{+}A$ can generate data for $A{+}B$ and $B{+}A$ without separately teleoperating each target pair. Here, $A$ and $B$ denote hand types ordered from left to right.

We consider two identical robot arms equipped with different combinations of hands. Collecting demonstrations separately for every pair presents two challenges. First, operators using heterogeneous pairs must coordinate both arms while accommodating differences in control mappings, mounting geometries, and contact behavior. Second, effective teleoperation of individual pairs does not eliminate the collection burden across combinations: $N$ hand types available on each arm yield $N^2$ ordered pairs. Generating demonstrations for multiple target pairs from a few source pairs offers an alternative to repeated manual collection.

Existing data-generation methods expand small sets of seed demonstrations into larger datasets~\cite{mandlekar2023mimicgen,xue2025demogen}, while cross-end-effector methods support demonstration reuse and policy learning through action alignment and shared latent spaces~\cite{zheng2025uniact,bauer2025latentdiff,mu2026one,jiang2026cross,wu2026cei}. However, the joint problem of executable demonstration generation and unified policy learning for bimanual systems with morphologically heterogeneous end-effector pairs remains largely underexplored.

A shared representation of hand states can support both objectives. Hands with different joint structures can realize similar interactions through different joint configurations, with correspondences in interaction location, approach direction, and finger spacing relative to object size. Encoding these correspondences in a common latent space supports transfer and provides a consistent action representation for a single policy trained across hand pairs.

We introduce AsymEEF, a framework that integrates cross-end-effector demonstration generation with unified bimanual policy learning, as illustrated in Figure~\ref{fig:teaser}. The Shared EEF-State Encoder maps the states of different hands into a shared latent space. Cross-end-effector decoding then produces candidate trajectories for target hand pairs, while quality checks and task-replay validation identify usable demonstrations. The same representation supports a unified latent bimanual visuomotor policy that jointly predicts latent actions for both arms from visual observations and robot states. These predictions are decoded into commands for the corresponding end effectors. We evaluate demonstration generation and policy learning on contact-rich simulated and real-world tasks. Our contributions are as follows:

\pagebreak
\begin{samepage}
\begin{itemize}
    \item We formulate executable cross-end-effector demonstration generation across homogeneous and heterogeneous bimanual pairs of two-, three-, and five-fingered hands.
    \item We introduce AsymEEF, combining cross-pair synthesis, quality checks, and task replay to generate validated simulated and real-world demonstrations.
    \item We develop a shared EEF-state representation for cross-hand transfer and unified latent bimanual policies.
    \item We evaluate generation, latent consistency, and policy performance on coordinated, sequential bimanual tasks.
\end{itemize}
\end{samepage}

\begin{table}[!htbp]
\caption{Comparison of Representative Related Work.}
\label{tab:related_work_positioning}
\centering
\small
\setlength{\tabcolsep}{4.2pt}
\renewcommand{\arraystretch}{1.12}

\providecommand{\cmark}{\ding{51}}
\providecommand{\xmark}{\ding{55}}
\providecolor{OursRow}{RGB}{240,240,240}

\begin{adjustbox}{max width=\linewidth}
\begin{tabular}{@{\hspace{2pt}}l l c l l@{\hspace{2pt}}}
\toprule
\multirow{2}{*}{\textbf{Method}}
& \multirow{2}{*}{\textbf{Arm Setting}}
& \multicolumn{2}{c}{\textbf{EEF Configuration}}
& \multirow{2}{*}{\textbf{Main Output}} \\
\cmidrule(lr){3-4}
& & \textbf{Cross-EEF} & \textbf{EEF Types} & \\
\midrule

\multicolumn{5}{@{\hspace{2pt}}l}{\textit{Demonstration generation}} \\
\addlinespace[1pt]

MimicGen~\cite{mandlekar2023mimicgen}
& Single & \xmark & Gripper & Synthetic demos \\

DemoGen~\cite{xue2025demogen}
& Single/dual & \xmark & Gripper, hand & Synthetic demos \\

DexMimicGen~\cite{jiang2025dexmimicgen}
& Dual & \xmark & Gripper, hand & Synthetic demos \\

MoMaGen~\cite{li2025momagen}
& Mobile dual & \xmark & Gripper & Synthetic demos \\

BiDemoSyn~\cite{zhou2025bidemosyn}
& Dual & \xmark & Gripper & Synthetic demos \\


CEI~\cite{wu2026cei}
& Single & \cmark & Gripper, hand & Synthetic demos \\

\midrule

\multicolumn{5}{@{\hspace{2pt}}l}{%
    \textit{Cross-embodiment / end-effector alignment}} \\
\addlinespace[1pt]

LAD~\cite{bauer2025latentdiff}
& Single & \cmark & Gripper, hand & Shared policy \\

XL-VLA~\cite{jiang2026cross}
& Dual & \cmark & Hand & Shared policy \\

OPFA~\cite{mu2026one}
& Single & \cmark & Gripper, hand & Shared policy \\

\midrule
\rowcolor{OursRow}
\textbf{Ours}
& Dual
& \cmark
& Gripper, hand
& \begin{tabular}[c]{@{}l@{}}
    Synthetic demos \\
    + shared policy
  \end{tabular}
\\
\bottomrule
\end{tabular}
\end{adjustbox}

\end{table}

\section{Related Work}

  \noindent
  \textbf{Bimanual manipulation and asymmetric coordination.}
  Bimanual manipulation expands the workspace and enables dexterous tasks, but requires coordinated control~\cite{smith2012dual}.
  Multi-arm methods coordinate motion through coupled planning, constraints, and task-and-motion
  formulations~\cite{doi:10.1177/0278364918765952,hartmann_long-horizon_2023}.
  Learning-based approaches acquire coordinated motions from demonstrations~\cite{liu2022stirr} and collaborative visual affordances from
  interaction data~\cite{zhao2023dualafford}.
  Bimanual teleoperation platforms support demonstration collection and policy learning~\cite{zhao2023learning,fu2024mobile}.
  Generalist and bimanual policies scale training with multi-robot or human
  data~\cite{openx2023,octo2024,black2024pi0,kim2024openvla,liu2024rdt}.
  Role-asymmetric methods assign persistent or phase-dependent stabilizing, supporting, acting, and handover
  roles~\cite{grannen2023stabilize,liu2024voxactb,yang2024asymdex,li2023handover}.
  These works address coordinated execution and policy learning rather than target-native demonstration generation for new end-effector
  pairs.

  \noindent
  \textbf{Demonstration generation from limited seeds.}
  Data-generation methods address demonstration scarcity.
  MimicGen adapts motion segments to new scene configurations, whereas DemoGen spatially augments trajectories and
  observations~\cite{mandlekar2023mimicgen,xue2025demogen}.
  This paradigm has expanded to bimanual dexterous and humanoid manipulation, mobile bimanual tasks, one-shot real-world synthesis, and
  video-diffusion augmentation through DexMimicGen, HumanoidGen, MoMaGen, BiDemoSyn, and
  CRAFT~\cite{jiang2025dexmimicgen,jing2025humanoidgen,li2025momagen,zhou2025bidemosyn,chen2026craft}.
  Most methods vary object poses, layouts, task instances, or appearance within a fixed control interface; DemoGen uses a 6-DoF
  abstraction, while CRAFT additionally supports video-and-action augmentation across embodiments.
  AsymEEF instead expands along the hardware axis by reconstructing target-native trajectories for heterogeneous
  bimanual end-effector pairs and retaining only trajectories that pass task replay.

  \noindent
  \textbf{Cross-embodiment and cross-end-effector alignment.}
  Cross-embodiment learning seeks reusable representations across heterogeneous robot action spaces.
  Large multi-robot datasets and generalist VLAs enable shared training but must reconcile embodiment-specific observations and
  actions~\cite{openx2023,octo2024,black2024pi0,kim2024openvla,zheng2025xvla,hou2025dita,tan2025blm1}.
  More directly, UniAct, LAD, OPFA, XL-VLA, CrossDex, and CEI introduce universal or latent action spaces, embodiment-conditioned
  decoders, retargeting mappings, or synthesized observations and actions across grippers and dexterous
  hands~\cite{zheng2025uniact,bauer2025latentdiff,mu2026one,jiang2026cross,yuan2025cross,wu2026cei}.
  Collectively, they establish the feasibility of cross-EEF sharing. Most emphasize policy co-training or adaptation, whereas CEI
  synthesizes cross-embodiment data in a single-arm setting.
  As summarized in Table~\ref{tab:related_work_positioning}, the joint problem of executable demonstration generation and unified policy
  learning for bimanual systems with morphologically heterogeneous end-effector pairs has received comparatively limited attention.
\section{Method}
\label{sec:method}

\subsection{Problem Formulation and Overview}
\label{sec:method_overview}

We consider bimanual systems composed of two identical arms equipped with an
ordered end-effector pair $b=(e^L,e^R)$. Given a small set of successful
demonstrations $\mathcal D_{b_s}$ collected with a source pair $b_s$, our goal
is to generate executable demonstrations $\widehat{\mathcal D}_{b_t}$ for a
target pair $b_t$ without teleoperating each target combination. We assume that
the target EEFs can perform the required interactions, although differences in
their actuation spaces and kinematic structures prevent direct reuse of joint
trajectories.

We represent an EEF state as $x_e=(T_e,\mathbf q_e)$, where
$T_e\in\mathrm{SE}(3)$ is the world-frame pose of its functional wrist frame
and $\mathbf q_e\in[-1,1]^{J_e}$ contains its normalized independent actuator
values. AsymEEF separates cross-EEF state alignment from trajectory-level
validation. Kinematic teacher anchors and a few real-world calibration anchors
supervise a shared encoder--decoder. This interface converts source trajectories
to target EEF pairs, while full-task replay determines which trajectories are
retained. The resulting demonstrations support both pair-specific policy
training and unified learning across pairs.

\subsection{Cross-EEF Anchoring}
\label{sec:cross_eef_anchoring}

\subsubsection{Geometry-Guided Teacher Anchoring}

For each EEF, we annotate functional contact points on URDF links.
Differentiable forward kinematics maps $\mathbf q_e$ to their positions
$\mathbf p_i(\mathbf q_e)$. We express these points in the functional wrist
frame, adopt consistent left/right conventions, and normalize lengths by an
EEF-specific characteristic scale.

The points are divided into two opposing groups with weighted centers
$\boldsymbol\mu_a$ and $\boldsymbol\mu_b$. We define the grasp center, opening
width, and signed opposition direction as
\begin{equation}
\begin{adjustbox}{max width=0.94\linewidth}$\displaystyle
\mathbf c_e=\frac{\boldsymbol\mu_a+\boldsymbol\mu_b}{2},\quad
w_e=\lVert\boldsymbol\mu_a-\boldsymbol\mu_b\rVert_2,\quad
\mathbf g_e=\frac{\boldsymbol\mu_a-\boldsymbol\mu_b}{w_e+\epsilon}.
$\end{adjustbox}
\label{eq:functional_geometry}
\end{equation}
For multifingered hands, the approach direction is the signed normal of a plane
defined by three fingertips in a fixed semantic order. For two-fingered
grippers, it points from the EEF base toward the grasp center. Thumb-tip
positions, thumb-pad normals, and named finger-chain keypoints provide
additional descriptors where available. Together, these quantities form a
functional geometry descriptor $\phi_e(\mathbf q_e)$ without requiring
identical joint layouts.

For each directed pair $e_s\rightarrow e_t$, we use scrambled Sobol sampling on
a morphology-aware grasp manifold and reject source states with invalid
functional frames. Given an admissible $\mathbf q_s$, a kinematic teacher
optimizes the target actuator state and a target-to-source functional-frame
transform $\Delta T$:
\begin{equation}
\begin{adjustbox}{max width=0.94\linewidth}$\displaystyle
\begin{aligned}
(\mathbf q_t^*,\Delta T^*)
&=\operatorname*{arg\,min}_{\mathbf q_t,\Delta T}
    \bigl(
    \mathcal L_{\mathrm{base}}
    +\mathcal L_{\mathrm{pair}}^{\rho(e_s,e_t)}
    +\lambda_w\mathcal L_{\mathrm{wrist}}
    \bigr),\\
\text{s.t.}\qquad
\mathbf q_t&\in[-1,1]^{J_t},
\qquad \Delta T\in\mathrm{SE}(3).
\end{aligned}
$\end{adjustbox}
\label{eq:teacher_anchor}
\end{equation}
Here, $\mathcal L_{\mathrm{base}}$ aligns grasp centers and signed gap and
approach directions, while $\mathcal L_{\mathrm{pair}}^{\rho}$ applies the
contact constraints appropriate to pair type $\rho$: effective aperture and
active contacts, thumb--opposed-contact relations, or multifinger-chain
geometry. When used, the wrist penalty discourages pose changes during in-hand
refinement, favoring compensation through target actuators; some pair types
instead fix selected pose degrees of freedom. We solve from multiple target
initializations and retain the best distinct candidates passing geometric,
directional, and physical-gap gates. Defining $\Delta T$ to map target-frame
coordinates into the source functional frame gives $T_t=T_s\Delta T^*$ for a
world-frame source pose $T_s$. The procedure requires registered kinematics,
actuator limits, and functional-point annotations, but no task demonstrations
or physical-robot measurements.

\begin{figure}[!htbp]
    \centering
    \includegraphics[width=0.78\linewidth]{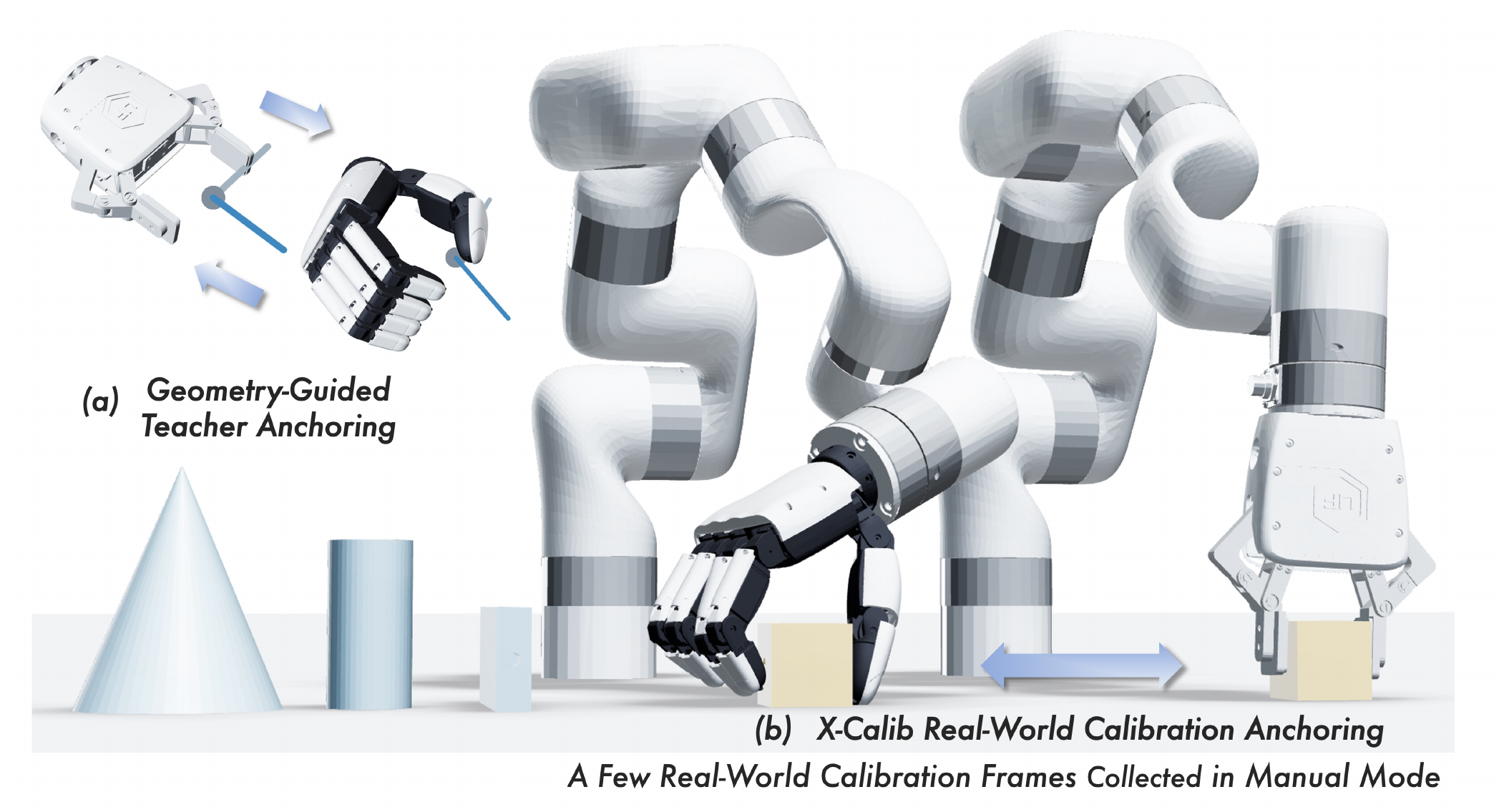}
    \caption{Geometry-guided teacher anchoring and X-Calib anchoring from a small set of real-world calibration frames.}
    \label{fig:teacher_anchor}

\end{figure}

\begin{figure}[!htbp]
    \centering
    \includegraphics[width=\textwidth]{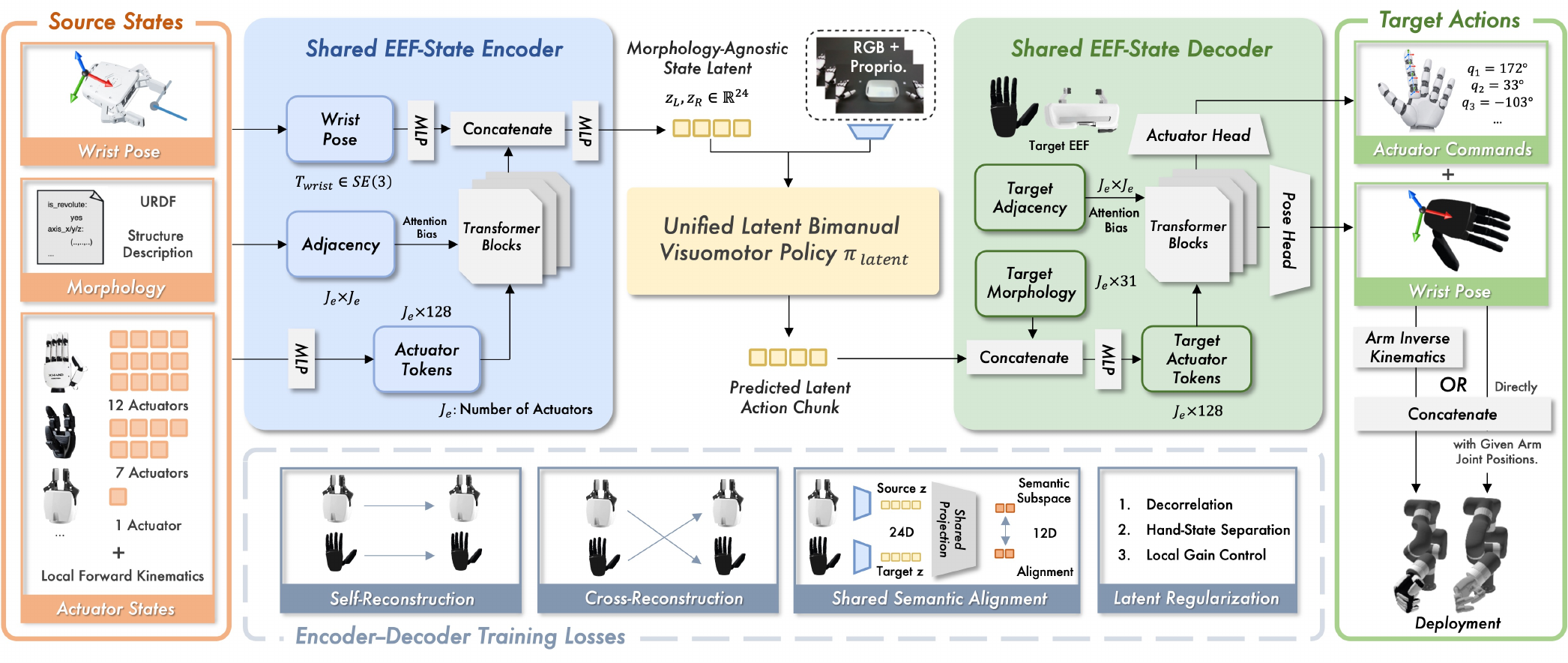}
    \caption{Architecture of AsymEEF. A shared encoder maps wrist poses, morphology, and actuator states into a morphology-agnostic latent space for bimanual visuomotor policy learning. The shared decoder produces target-specific actuator commands and wrist poses, with arm motion obtained through inverse kinematics or given joint positions. Training combines self-reconstruction, cross-reconstruction, semantic alignment, and latent regularization.}
    \label{fig:architecture}

\end{figure}

\subsubsection{X-Calib Real-World Calibration Anchoring}

Kinematic similarity does not capture assembly offsets, actuator coupling, or
contact compliance. X-Calib therefore adds real grasp keyframes for the
available physical EEFs. Each calibration intent
\begin{equation}
    \kappa=(o,\ell,r)
\label{eq:xcalib_intent}
\end{equation}
specifies an object, a discrete workspace location, and a grasp orientation.
Stable arm and hand configurations are recorded through manual teaching.
Keyframes sharing $\kappa$ form weak correspondences. Each anchor is represented
as $(x_s,\mathcal Y_t)$, where $\mathcal Y_t$ contains valid target states and
preserves alternative configurations for the same grasp intent. Functional
wrist poses are recovered through full-robot forward kinematics; arm joints are
used only for pose recovery and are excluded from the EEF latent. Teacher
anchors provide broad spatial coverage, while X-Calib captures
hardware-specific deviations, as shown in Figure~\ref{fig:teacher_anchor}.

\subsection{Shared EEF-State Interface}
\label{sec:shared_eef_interface}

\subsubsection{Encoder--Decoder Architecture}

The shared interface maps variable-dimensional EEF states to a fixed-dimensional
latent and decodes them according to the target morphology:
\begin{equation}
\begin{aligned}
    \mathbf z_e
    &=E_\theta(x_e;\mathbf M_e,\mathbf A_e),\\
    \widehat x_t
    &=D_\psi(\mathbf z_e;\mathbf M_t,\mathbf A_t),
\end{aligned}
\label{eq:eef_interface}
\end{equation}
where $\mathbf M_e$ and $\mathbf A_e$ describe URDF-derived actuator morphology
and kinematic adjacency. EEF identity only selects these structural descriptors;
it does not select separate network parameters.

The \emph{Shared EEF-State Encoder} constructs one token per independent
actuator. Each token combines its current value, local FK position relative to
the functional wrist, and static features describing joint type, axis, mounting
transform, range, and mimic relationships. A topology-biased Transformer~\cite{vaswani2017attention}
processes these tokens together with a learned state token. Kinematic adjacency
biases attention toward neighboring joints while retaining global interactions
between fingers.

Wrist translation and a continuous six-dimensional rotation representation~\cite{zhou2019continuity} are
embedded separately and fused with the state-token output to produce a
24-dimensional latent. The decoder conditions target-morphology actuator
queries and a wrist query on this latent. Shared actuator and pose heads then
recover the target actuator state and functional wrist pose. Our implementation
uses 128-dimensional tokens, three Transformer layers, four attention heads,
and supports up to 12 independent actuators. The latent jointly represents
wrist pose and internal hand configuration without manually assigned semantic
blocks. Figure~\ref{fig:architecture} summarizes the architecture and its
connection to the latent bimanual policy.

\subsubsection{Joint Training Objective}

We jointly train one encoder and one decoder on both anchor sources:
\begin{equation}
\begin{aligned}
\mathcal L_{\mathrm{int}}
={}&\lambda_{\mathrm{self}}\mathcal L_{\mathrm{self}}
 +\lambda_{\mathrm{cross}}\mathcal L_{\mathrm{cross}}\\
 &+\lambda_{\mathrm{align}}\mathcal L_{\mathrm{align}}
 +\lambda_{\mathrm{reg}}\mathcal L_{\mathrm{reg}}.
\end{aligned}
\label{eq:interface_objective}
\end{equation}
For brevity, $E_e$ and $D_e$ denote conditioning on
$(\mathbf M_e,\mathbf A_e)$. Let $\ell_e(\widehat x,x)$ combine actuator-state
error, wrist-translation error, and rotational geodesic error. Self- and
cross-reconstruction are defined as
\begin{equation}
\begin{aligned}
\mathcal L_{\mathrm{self}}
&=\mathbb E_{e,x_e}
  \left[\ell_e\!\left(D_e(E_e(x_e)),x_e\right)\right],\\
\mathcal L_{\mathrm{cross}}
&=\mathbb E_{(x_s,\mathcal Y_t)}
  \left[
  \ell_t^{\mathrm{set}}
  \!\left(D_t(E_s(x_s)),\mathcal Y_t\right)
  \right].
\end{aligned}
\label{eq:reconstruction_losses}
\end{equation}
The set loss $\ell_t^{\mathrm{set}}$ applies soft selection among valid target
states according to their combined actuator and wrist errors, rather than
constructing an averaged target configuration. Teacher anchors correspond to
singleton target sets.

Requiring complete latent equality would suppress configurations that a
dexterous hand can express but a gripper cannot. We therefore align only a
training-time projection $P:\mathbb R^{24}\rightarrow\mathbb R^{12}$:
\begin{equation}
\mathcal L_{\mathrm{align}}
=
\mathbb E_{(x_s,x_t)}
\left[
1-\cos\!\left(P(E_s(x_s)),P(E_t(x_t))\right)
\right].
\label{eq:latent_alignment}
\end{equation}
The expectation is taken over corresponding anchor states. The regularization
term maintains latent variance, reduces redundancy, preserves sensitivity to
changes in internal hand configuration, and limits local amplification in
cross-EEF mappings. Subsequent generation and policy learning use the full
24-dimensional latent rather than the alignment projection.

\subsection{Demonstration Generation and Validation}
\label{sec:generation_validation}

\subsubsection{Cross-Pair Transfer and Replay}

We map each EEF state on the shared source timeline as follows:
\begin{equation}
    \widehat x_{t,k}^{\,i}
    =
    D_{e_t^i}\!\left(E_{e_s^i}(x_{s,k}^{\,i})\right),
    \qquad i\in\{L,R\},
\label{eq:cross_pair_transfer}
\end{equation}
where $k$ indexes trajectory steps. Decoded wrist poses are converted to arm
commands through constrained inverse kinematics, while decoded actuator states
control the target EEFs. The complete task is then replayed through simulation
physics or on the target physical robot.

\begin{figure}[!htbp]
    \centering
    \includegraphics[width=0.80\linewidth]{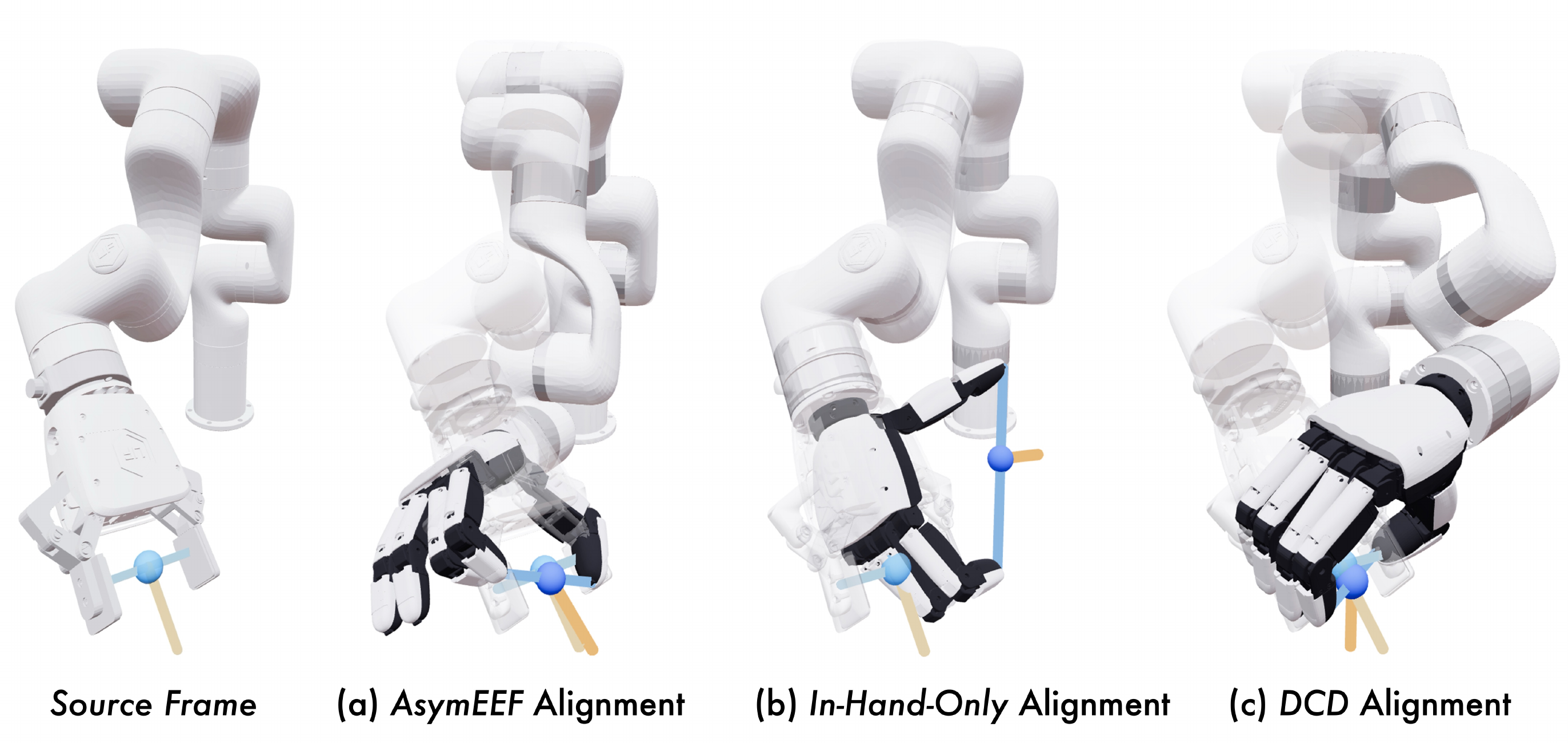}
    \caption{Target EEF alignment relative to the source functional grasp frame.}
    \label{fig:misalign}

\end{figure}

Locally aligned states do not guarantee successful interactions over an entire
trajectory. We therefore retain a candidate only if it satisfies kinematic
feasibility, joint continuity, collision constraints, valid contact sequences,
and task success. Intended object contacts are permitted; unintended collisions
are rejected. We neither overwrite object states during replay nor apply
external forces to rescue failures. Accepted executions provide observations
and control sequences for the target pair. Thus, anchor consistency guides
transfer, but task execution determines whether a generated trajectory becomes
a usable demonstration. Figure~\ref{fig:misalign} compares the transferred
target states with the source functional grasp frame and alternative alignment
strategies.
For tasks where objects impose tight grasp-orientation constraints beyond simple pick-and-place, both alternatives can fail when mounting-induced wrist-frame offsets are not adequately accounted for. We discuss these failure modes in Sec.~\ref{sec:experiments}.

\subsubsection{Progressive Spatial Expansion}

To increase simulation coverage, we divide successful seeds into approach,
grasp, and manipulation segments and remap wrist trajectories to randomized
object configurations. Expansion starts with small perturbations around the
seeds and gradually covers larger object translations, robot-base
perturbations, and workspace boundaries. Arm trajectories are recomputed with
IK, while EEF states are retained or transferred through the shared interface.
Every candidate undergoes the same replay validation, and successful
trajectories can seed subsequent rounds. This procedure concentrates
generation on regions that continue to produce valid demonstrations, rather
than sampling a broad configuration space indiscriminately.

\subsection{Unified Latent Bimanual Visuomotor Policy}
\label{sec:unified_policy}

We freeze the encoder and decoder and encode all demonstrations offline. For the
seven-joint arms in our system, the proprioceptive state is
\begin{equation}
\mathbf s_k=
\left[
\mathbf q_k^{\mathrm{arm},L},
\mathbf z_k^L,
\mathbf q_k^{\mathrm{arm},R},
\mathbf z_k^R
\right]
\in\mathbb R^{62}.
\label{eq:policy_state}
\end{equation}
Actions use the same layout, but contain future arm-joint targets and latent
encodings of desired EEF states rather than current measurements. Native and
generated demonstrations from different pairs can therefore be mixed without
morphology-specific policy heads.

We evaluate ACT~\cite{zhao2023learning} and $\pi_{0.5}$~\cite{black2025pi05} as
representative backbones for the \emph{unified latent bimanual visuomotor
policy}. Both receive visual observations and the shared proprioceptive state
and predict action chunks through the same interface. Each retains its original
training objective: ACT is trained for action-chunk prediction, while
$\pi_{0.5}$ uses its native flow-matching objective. This separates the
cross-EEF representation from the choice of imitation learner.

At inference, the frozen encoder maps wrist poses and EEF feedback to latent observations. Arm actions are realized in either joint- or task-space mode. In the former, when arm qpos is included in the training action representation, the policy directly predicts arm-joint targets, while the target-conditioned decoder recovers EEF actuator commands. In the latter, decoded wrist targets are converted online to arm commands through IK. The modes are mutually exclusive, avoiding redundant arm commands while retaining fixed-dimensional observations and actions across EEF pairs. This design also provides a natural extension to cross-embodiment transfer across arm morphologies. Unless otherwise noted, all subsequent policy evaluations use the joint-space mode.
\section{Experiments}
\label{sec:experiments}

We evaluate whether AsymEEF generates executable cross-EEF demonstrations,
increases usable data yield, and supports downstream policy learning.
Representation ablations examine the contributions of the training objectives and real-world calibration.



\begin{table}[!htbp]

\caption{
Main Evaluation: Demonstration Replay and Policy Evaluation on Generated Data.
}
\label{tab:main_results}

\centering
\fontsize{9}{11}\selectfont

\setlength{\tabcolsep}{2.8pt}
\renewcommand{\arraystretch}{1.12}

\newcommand{\na}{--}
\newcommand{\resultgray}{gray!12}


\newcommand{\headstrut}{%
    \rule[-0.75ex]{0pt}{2.85ex}%
}

\newcommand{\metrichead}[1]{%
    \headstrut\textbf{#1}%
}

\newcommand{\grouphead}[1]{%
    \headstrut\textbf{#1}%
}

\newcommand{\metahead}[1]{%
    \multirow[c]{2}{=}{%
        \centering
        \raisebox{-0.30ex}{\textbf{#1}}%
    }%
}


\newcommand{\simsource}{%
    \begin{tabular}[c]{@{}l@{}}
        Franka + Franka \\
        Ability + Ability
    \end{tabular}%
}


\begin{adjustbox}{max width=\linewidth}
\begin{tabular}{
    >{\centering\arraybackslash}m{0.52in}
    >{\centering\arraybackslash}m{1.05in}
    >{\raggedright\arraybackslash}m{1.10in}
    >{\raggedright\arraybackslash}m{1.15in}
    >{\centering\arraybackslash}m{0.48in}
    >{\centering\arraybackslash}m{0.48in}
    >{\columncolor{\resultgray}\centering\arraybackslash\bfseries}m{0.48in}
    >{\columncolor{\resultgray}\centering\arraybackslash}m{0.48in}
    >{\columncolor{\resultgray}\centering\arraybackslash}m{0.48in}
}

\toprule


\metahead{Domain}
&
\metahead{Task}
&
\metahead{Source EEF Pair}
&
\metahead{Target EEF Pair}
&
\multicolumn{3}{c}{
    \grouphead{Data Gen. (Replay SR)}\,$\uparrow$
}
&
\multicolumn{2}{c}{
    \grouphead{Policy Eval. (SR)}\,$\uparrow$
}
\\

\cmidrule(lr){5-7}
\cmidrule(lr){8-9}


& & & &
\metrichead{LAD}
&
\metrichead{CEI}
&
\metrichead{Ours}
&
\metrichead{ACT}
&
\headstrut$\boldsymbol{\pi}_{0.5}$
\\

\midrule


\multirow{12}{*}{Sim}


&
\multirow{3}{*}{\textit{Can Sorting}}
&
\multirow{3}{*}{Franka + Franka}
&
Franka + Ability
&
0.14
&
0.10
&
0.57
&
0.91
&
1.00
\\

& & &
Franka + XHAND
&
0.14
&
0.19
&
0.62
&
1.00
&
1.00
\\

& & &
Franka + Dex3-1
&
0.38
&
0.48
&
0.86
&
1.00
&
1.00
\\

\cmidrule(lr){2-9}


&
\multirow{3}{*}{\textit{Handover}}
&
\multirow{3}{*}{Franka + Franka}
&
Franka + Ability
&
0.10
&
0.10
&
0.71
&
0.73
&
0.91
\\

& & &
Franka + XHAND
&
0.19
&
0.14
&
0.52
&
0.82
&
0.91
\\

& & &
Franka + Dex3-1
&
0.38
&
0.24
&
0.95
&
1.00
&
1.00
\\

\cmidrule(lr){2-9}


&
\multirow{3}{*}{\textit{Lift Tray}}
&
\multirow{3}{*}{Franka + Franka}
&
Franka + Ability
&
0.33
&
0.43
&
0.52
&
0.73
&
0.73
\\

& & &
Franka + XHAND
&
0.24
&
0.48
&
0.71
&
0.91
&
0.91
\\

& & &
Franka + Dex3-1
&
0.29
&
0.24
&
0.57
&
0.64
&
0.73
\\

\cmidrule(lr){2-9}


&
\multirow{3}{*}{\textit{Laptop \& Mouse}}
&
\multirow{3}{*}{Ability + Ability}
&
Ability + Franka
&
0.58
&
0.33
&
0.75
&
0.82
&
1.00
\\

& & &
Ability + XHAND
&
0.50
&
0.33
&
0.83
&
1.00
&
0.91
\\

& & &
Ability + Dex3-1
&
0.50
&
0.42
&
0.67
&
0.82
&
0.73
\\

\midrule


\multirow{6}{*}{Real}


&
\textit{Toy Sorting}
&
XHAND + XHAND
&
XHAND + G2
&
0.29
&
0.82
&
0.94
&
1.00
&
1.00
\\


&
\textit{Handover}
&
G2 + G2
&
XHAND + G2
&
0.32
&
0.44
&
0.79
&
0.91
&
1.00
\\


&
\textit{Motherboard}
&
XHAND + XHAND
&
XHAND + G2
&
0.20
&
0.44
&
0.84
&
0.91
&
0.82
\\


&
\textit{CPU Installation}
&
G2 + G2
&
XHAND + G2
&
0.31
&
0.00
&
0.79
&
0.55
&
0.73
\\


&
\textit{Screwdriver Bit}
&
G2 (single)
&
Bracket + G2
&
0.00
&
0.71
&
1.00
&
1.00
&
0.82
\\


&
\textit{Laptop \& Mouse}
&
G2 + G2
&
XHAND + G2
&
0.13
&
0.30
&
0.96
&
0.64
&
0.64
\\

\bottomrule

\end{tabular}
\end{adjustbox}

\end{table}
\begin{figure}[!htbp]

    \centering
    \includegraphics[width=\textwidth]{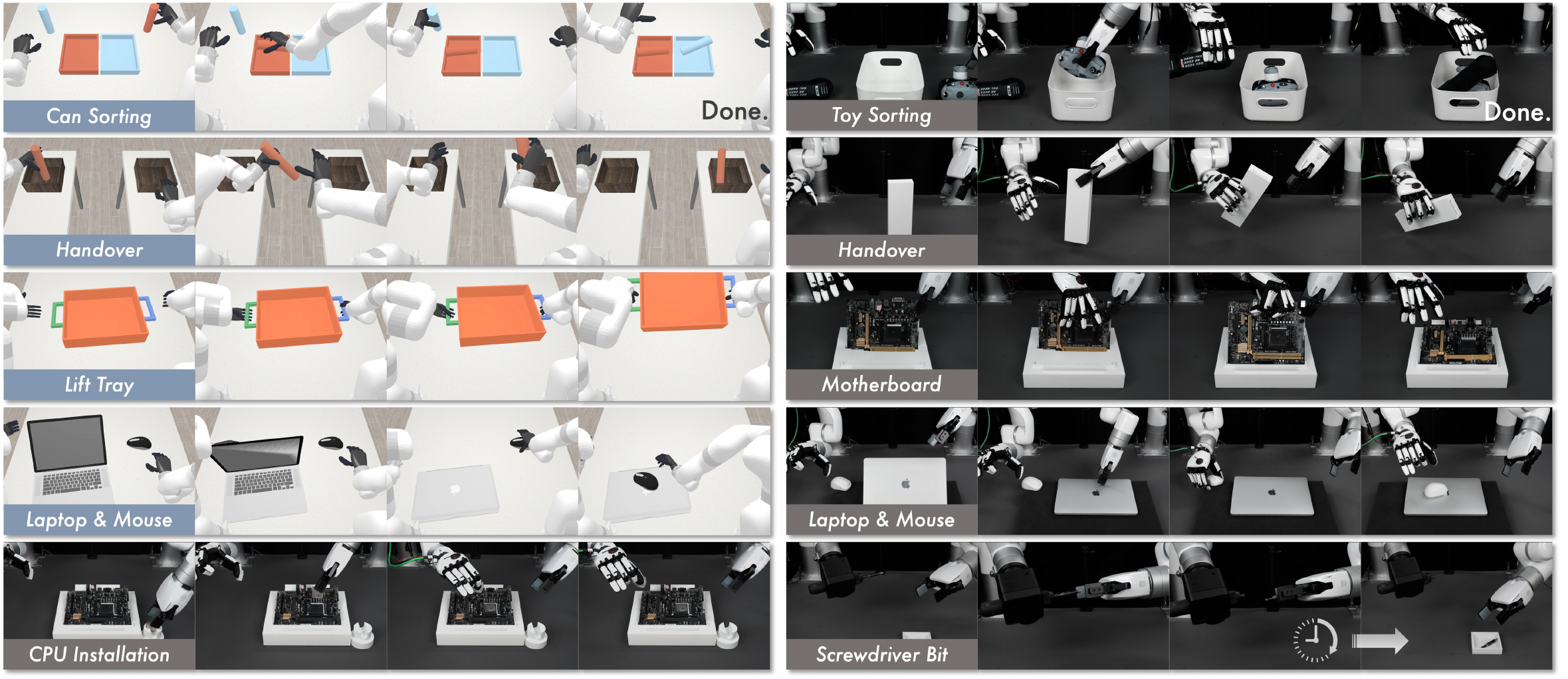}
    \caption{Full task sequences in SAPIEN simulation and on real robots, including bimanual collaboration and tool manipulation.}
    \label{fig:tasks}

\end{figure}


\subsection{Experimental Setup}

\paragraph{Platform and tasks}
Our real system consists of xArm7 arms, with XHANDs, G2 grippers, and a screwdriver mounting bracket. Teleoperation uses a Meta Quest 3 with
controllers and Wuji gloves. Models are trained on RTX PRO 5000 GPUs. Across both domains, we evaluate
XHAND, Ability Hand, Unitree Dex3-1, Franka hand, and G2 gripper.

We use four simulated tasks---\textit{Can Sorting}, \textit{Handover},
\textit{Lift Tray}, and \textit{Laptop \& Mouse}---and six real tasks---
\textit{Toy Sorting}, \textit{Handover}, \textit{Motherboard}, \textit{CPU
Installation}, \textit{Screwdriver Bit}, and \textit{Laptop \& Mouse}. These cover
sequential manipulation, coordination, precise insertion, and
orientation-sensitive interaction. In \textit{Screwdriver Bit}, one arm provides fixed tool
support while the other hand removes the detachable bit. Figure~\ref{fig:tasks} shows full task
sequences of our experiment.

\paragraph{Metrics and protocol}
Replay SR measures the fraction of transferred candidates that complete the
full task on the target EEF pair. Policy SR measures task completion in
closed-loop policy rollouts. We define replay-valid expansion gain and candidate
acceptance rate as
\begin{equation}
G_{\mathrm{exp}}=\frac{N_{\mathrm{acc}}}{N_{\mathrm{src}}},
\qquad
R_{\mathrm{acc}}=\frac{N_{\mathrm{acc}}}{N_{\mathrm{gen}}},
\label{eq:expansion_metrics}
\end{equation}
where $N_{\mathrm{src}}$, $N_{\mathrm{gen}}$, and $N_{\mathrm{acc}}$ count
teleoperated source demonstrations, generated candidates, and replay-accepted
trajectories, respectively. Generated trajectories reused as seeds do not
increase $N_{\mathrm{src}}$. For the direct replay evaluation in Table~II, each method is evaluated on 12-21 candidates per simulated setting and 17-34 candidates per real-world setting.

Within each setting, all methods use the same source demonstrations and matched initial object poses and robot configurations. All methods follow the same replay protocol,
requiring kinematic feasibility, joint and collision validity, valid contact
progression, and sustained task success. Object states are not overwritten
after initialization, and failed executions receive no external assistance.

\paragraph{X-Calib collection}
We collect calibration frames at three fixed workspace positions with four
3D-printed geometries: a rectangular cuboid, cube, cylinder, and cone.
At each location, we sample two or four prescribed grasp orientations and
record three static frames per selected pose. Collection requires only one arm,
as the opposite side is obtained by mirroring. The final XHAND--G2 calibration
set contains 216 frames and requires less than 1 hour to collect.

\begin{table}[!htbp]
\caption[Cross-End-Effector Data Expansion in SAPIEN.]{%
\textsc{Cross-End-Effector Data Expansion in SAPIEN.}%
}
\label{tab:sim_data_expansion}
\centering
\small
\setlength{\tabcolsep}{5.0pt}
\renewcommand{\arraystretch}{1.08}
\newcommand{\seef}[2]{#1}
\newcommand{\eef}[2]{#1}
\newcommand{\best}[1]{\textbf{#1}}

\begin{adjustbox}{max width=\linewidth}
\begin{tabular}{lllccccc}
\toprule
\textbf{Task} & \textbf{Source} & \textbf{Target}
& $\mathbf{N}_{\mathrm{src}}$ & $\mathbf{N}_{\mathrm{gen}}$
& $\mathbf{N}_{\mathrm{acc}}$ & \textbf{Acc.} & \textbf{Gain}\, $\uparrow$ \\
\midrule
\multirow{6}{*}{\textit{Can Sorting}}
& \multirow{3}{*}{\seef{Franka}{2}} & \eef{Dex3-1}{3}
& \multirow{3}{*}{20} & 400 & 296 & 0.74 & 14.80$\times$ \\
& & \eef{XHAND}{5} & & 494 & 322 & 0.65 & \best{16.10$\times$} \\
& & \eef{Ability}{5} & & 407 & 317 & \best{0.78} & 15.85$\times$ \\
\cmidrule(lr){2-8}
& \multirow{3}{*}{\seef{Ability}{5}} & \eef{Franka}{2}
& \multirow{3}{*}{21} & 504 & 333 & 0.66 & \best{15.86$\times$} \\
& & \eef{XHAND}{5} & & 432 & 317 & \best{0.73} & 15.10$\times$ \\
& & \eef{Dex3-1}{3} & & 462 & 331 & 0.72 & 15.76$\times$ \\
\midrule
\multirow{6}{*}{\textit{Handover}}
& \multirow{3}{*}{\seef{Franka}{2}} & \eef{Dex3-1}{3}
& \multirow{3}{*}{20} & 400 & 304 & \best{0.76} & 15.20$\times$ \\
& & \eef{XHAND}{5} & & 494 & 319 & 0.65 & 15.95$\times$ \\
& & \eef{Ability}{5} & & 494 & 325 & 0.66 & \best{16.25$\times$} \\
\cmidrule(lr){2-8}
& \multirow{3}{*}{\seef{Ability}{5}} & \eef{Franka}{2}
& \multirow{3}{*}{11} & 202 & 202 & \best{1.00} & \best{18.36$\times$} \\
& & \eef{XHAND}{5} & & 286 & 182 & 0.64 & 16.55$\times$ \\
& & \eef{Dex3-1}{3} & & 294 & 125 & 0.43 & 11.36$\times$ \\
\midrule
\multirow{6}{*}{\textit{Lift Tray}}
& \multirow{3}{*}{\seef{Franka}{2}} & \eef{Dex3-1}{3}
& \multirow{3}{*}{20} & 400 & 385 & 0.96 & 19.25$\times$ \\
& & \eef{XHAND}{5} & & 400 & 378 & 0.95 & 18.90$\times$ \\
& & \eef{Ability}{5} & & 400 & 393 & \best{0.98} & \best{19.65$\times$} \\
\cmidrule(lr){2-8}
& \multirow{3}{*}{\seef{Ability}{5}} & \eef{Franka}{2}
& \multirow{3}{*}{8} & 160 & 124 & 0.78 & 15.50$\times$ \\
& & \eef{XHAND}{5} & & 160 & 147 & 0.92 & 18.38$\times$ \\
& & \eef{Dex3-1}{3} & & 160 & 159 & \best{0.99} & \best{19.88$\times$} \\
\bottomrule
\end{tabular}
\end{adjustbox}

\end{table}

\paragraph{Baseline Methods}
We compare two representative cross-EEF transfer methods:
LAD~\cite{bauer2025latentdiff} and CEI~\cite{wu2026cei}. LAD transfers latent actions that capture in-hand pose
semantics but does not explicitly account for arm- and wrist-pose discrepancies
induced by EEF mounting. This limitation is critical in orientation-sensitive
tasks such as \textit{Motherboard} and \textit{Handover}, where successful
interaction requires precise approach directions and object orientations.
CEI uses Directional Chamfer Distance to align fingertip surface-normal
sets across EEFs. It constrains the pinch direction but leaves the wrist
approach orientation underconstrained. Figure~\ref{fig:misalign} illustrates
their respective misalignment modes as \emph{In-Hand-Only} and \emph{DCD}.
Both methods are evaluated using the same replay protocol as AsymEEF.

\subsection{Cross-EEF Generation and Data Expansion}

Table~II reports direct replay results.
Across all 12 simulated task--EEF settings and six real-world tasks,
AsymEEF achieves higher Replay SR than both LAD and CEI. Across six real tasks, the
corresponding means are $88.7\%$, $20.8\%$, and $45.2\%$. On CPU Installation,
which requires precise placement and orientation, AsymEEF reaches $79.0\%$,
versus $31.0\%$ and $0.0\%$. These results demonstrate higher task-level
transfer success under the same criteria.
We next evaluate progressive spatial expansion on three simulated tasks
(Table~\ref{tab:sim_data_expansion}). Across 18 source--target settings, 4,959
of 6,549 candidates pass replay, yielding an aggregate acceptance rate of
$75.7\%$. Each source demonstration produces $11.36$--$19.88$ accepted
trajectories, with a mean expansion gain of $16.6\times$ across settings. These
gains count successful executions rather than geometric transformations. The
acceptance rate applies specifically to progressive expansion on the selected
tasks and is distinct from the replay evaluation in
Table~\ref{tab:main_results}.

\subsection{Downstream Policy Learning}

We train the unified latent bimanual visuomotor policy on replay-filtered
demonstrations using ACT and $\pi_{0.5}$ as backbones. The Shared EEF-State
Encoder and decoder remain frozen. For each task, policy training uses replay-accepted demonstrations from each target EEF pair and native source demonstrations. For each task and backbone, we train a single policy jointly across the included EEF pairs. Both backbones retain their standard
imitation-learning objectives, with matched data splits, visual inputs, and
evaluation initializations.
Across six real tasks, ACT and $\pi_{0.5}$ both achieve a mean Policy SR of
$83.3\%$, with each task executed 11 times. ACT ranges from $55\%$ to $100\%$, reaching $100\%$ on \textit{Toy
Sorting} and \textit{Screwdriver Bit} and $91\%$ on \textit{Handover} and
\textit{Motherboard}. Using the same latent interface, $\pi_{0.5}$ ranges from
$64\%$ to $100\%$, reaching $100\%$ on \textit{Toy Sorting} and
\textit{Handover}, $82\%$ on \textit{Motherboard} and \textit{Screwdriver Bit},
$73\%$ on \textit{CPU Installation}, and $64\%$ on \textit{Laptop \& Mouse}.
These results show that replay-filtered demonstrations provide useful
supervision for closed-loop policy learning, beyond direct trajectory replay.

\subsection{Representation and Anchoring Ablations}

Table~\ref{tab:eef_ablation} evaluates within-EEF reconstruction (Native MAE),
cross-EEF decoding (Cross MAE), and cross-EEF reconstruction on real hardware
data (Real Cross MAE). The Avg. normalized error in Table IV is computed by averaging the normalized Native MAE, Cross MAE, wrist translation error, wrist rotation error, and Real Cross MAE. The full model achieves the lowest average normalized
error of $4.185$. Removing
self-reconstruction reduces Cross MAE from $0.3069$ to $0.2504$ and Real Cross
MAE from $0.2088$ to $0.1766$, but increases Native MAE from $0.0967$ to
$0.5617$, raising the average normalized error to $8.325$. Self-reconstruction
preserves within-EEF state information rather than uniformly improving transfer
accuracy. Removing cross-reconstruction leaves
Native MAE nearly unchanged but increases Cross MAE to $0.5813$ and Real Cross
MAE to $0.3345$, supporting explicit cross-EEF supervision rather than relying
on within-EEF reconstruction alone.

Removing X-Calib leaves Native MAE nearly unchanged and slightly reduces Cross
MAE, yet increases Real Cross MAE from $0.2088$ to $0.6181$, nearly threefold.
This supports supplementing kinematic supervision with sparse physical
anchors to improve reconstruction on real hardware. Together, the ablations
distinguish the contributions of state preservation, cross-EEF supervision, and
physical calibration, with the full model achieving the best aggregate
performance across these objectives.

\begin{table}[H]
\caption{Ablation of EEF Representation Learning.}
\label{tab:eef_ablation}
\centering
\small
\setlength{\tabcolsep}{2.2pt}
\renewcommand{\arraystretch}{1.12}
\begin{tabular*}{\columnwidth}{@{\extracolsep{\fill}}lcccc}
\toprule
\textbf{Variant}
& \shortstack{\textbf{Avg. normalized}\\\textbf{error} $\downarrow$}
& \shortstack{\textbf{Native}\\\textbf{MAE} $\downarrow$}
& \shortstack{\textbf{Cross}\\\textbf{MAE} $\downarrow$}
& \shortstack{\textbf{Real Cross}\\\textbf{MAE} $\downarrow$} \\
\midrule
\textbf{Full}        & \textbf{4.185} & \textbf{0.0967} & 0.3069 & 0.2088 \\
\textit{w/o Self}    & \cellcolor{OursRow}8.325 & \cellcolor{OursRow}0.5617 & \textbf{0.2504} & \textbf{0.1766} \\
\textit{w/o Cross}   & 6.858 & 0.0973 & \cellcolor{OursRow}0.5813 & 0.3345 \\
\textit{w/o X-Calib} & 5.697 & 0.0992 & 0.2937 & \cellcolor{OursRow}0.6181 \\
\bottomrule
\end{tabular*}

\end{table}

\Needspace{10\baselineskip}
\section{Conclusion}
\label{sec:conclusion}

We presented AsymEEF, which connects cross-end-effector demonstration
generation and unified bimanual policy learning through a shared EEF-state
interface. Experiments demonstrate that transferring and validating
demonstrations across hand morphologies increases usable data yield and supports
closed-loop policy learning. These results support collecting demonstrations with a few source
pairs and reusing them across additional configurations. Our evaluation focuses
on tabletop manipulation, and X-Calib requires manually collected calibration
keyframes. Broader task coverage and automated calibration are natural
directions for extending the framework.


\pagebreak
\begingroup
\small
\setlength{\parskip}{0pt}
\bibliographystyle{IEEEtran}
\bibliography{references}
\par\pagebreak
\endgroup
\end{document}